\documentclass[journal]{IEEEtran}
\usepackage{amsfonts}
\usepackage{amssymb}
\usepackage{amsmath}
\usepackage{algorithm}
\usepackage{multirow}
\usepackage{xcolor}
\usepackage{graphicx}
\usepackage{cite}
\usepackage{exscale}
\usepackage{relsize}
\usepackage{algpseudocode}
\usepackage{graphics}
\ifCLASSOPTIONcompsoc
\usepackage[caption=false,font=normalsize,labelfont=sf,textfont=sf]{subfig}
\else
\usepackage[caption=false,font=footnotesize]{subfig}
\fi

\newcommand{\bm}[1]{\mbox{\boldmath{$#1$}}}

\usepackage[T1]{fontenc}

\begin{document}

\title{High-dimensional Metric Combining for Non-coherent Molecular Signal Detection}

\author{\IEEEauthorblockN{Zhuangkun~Wei,
                          Weisi~Guo, Bin~Li, Jerome Charmet, and Chenglin~Zhao}

\thanks{Zhuangkun~Wei, Weisi Guo and Jerome Charmet are with the University of Warwick,
Coventry, West Midlands, CV4~7AL, UK. (Email: Zhuangkun.wei@warwick.ac.uk)}
\thanks{Bin Li and Chenglin Zhao are with the School of Information and Communication Engineering (SICE),
Beijing University of Posts and Telecommunications (BUPT), Beijing,
100876, China.}
}

\maketitle

\begin{abstract}
In emerging Internet-of-Nano-Thing (IoNT), information will be embedded and conveyed in the form of molecules through complex and diffusive medias. One main challenge lies in the long-tail nature of the channel response causing inter-symbol-interference (ISI), which deteriorates the detection performance. If the channel is unknown, we cannot easily achieve traditional coherent channel estimation and cancellation, and the impact of ISI will be more severe. In this paper, we develop a novel high-dimensional non-coherent scheme for blind detection of molecular signals. We achieve this in a higher-dimensional metric space by combining different non-coherent metrics that exploit the transient features of the signals. By deducing the theoretical bit error rate (BER) for any constructed high-dimensional non-coherent metric, we prove that, higher dimensionality always achieves a lower BER in the same sample space. Then, we design a generalised blind detection algorithm that utilizes the Parzen approximation and its probabilistic neural network (Parzen-PNN) to detect information bits. Taking advantages of its fast convergence and parallel implementation, our proposed scheme can meet the needs of detection accuracy and real-time computing. Numerical simulations demonstrate that our proposed scheme can gain 10dB BER compared with other state of the art methods.
\end{abstract}


\IEEEpeerreviewmaketitle

\section{Introduction}
\IEEEPARstart{R}{ecent} advances in nano-technology have attracted widespread interests in a range of applications, including the Internet-of-Nano-Things (IoNT) \cite{Kahl2013, 7060516}, precision medicine \cite{7587347}, and covert signalling in electromagnetically denied environments. To support these, increasing efforts have been spent on the studies of molecular communications via diffusion (MCvD), whereby digital information is modulated via the chemical structures of molecules, and then undergo a combination of diffusion and advection propagation. Compared with the traditional electromagnetic waves (EMW) or acoustic waves \cite{7181710}, the stochastic nature of the propagation allows nano-scale information carriers to diffuse through complex medias, thereby making MCvD a promising communication candidate for bioengineering applications.

In the context of the signal detection in the MCvD, three challenges should be considered. First, the long-tail nature of the channel response causes severe inter-symbol-interference (ISI), which will deteriorate the accuracy of signal detection. Second, the underlying diffusion model will be inaccurate (or even unavailable) when dealing with complex channels (e.g., a micro-fluidic and absorbing channels with complex and irregular shapes obstacles \cite{8513142, 7839303, 6807659, 7073637, 6168841}). In this view, understanding a total channel state information (CSI) requires a large amounts of resources, therefore making those model-related detection schemes less practical. Third, a nano-receiver aiming at collecting the number of target molecules or measuring their concentration should be subject to a limited energy expenditure and also ensure real-time communication.

\subsection{Related Works}
The pure mass diffusion channel impulse response (CIR) and the noise distribution formulation are well established (even for certain boundary and reaction conditions). As such, there are significant efforts on developing coherent signal processing schemes leveraged on the computable CSIs. For example, various inference schemes aimed at jointly detecting the molecular information as well as estimating the underlying channel parameters have been proposed in \cite{6708551, 6868273, 7553035}. The state-of-the-art MAP \cite{6708551} uses a designed pilot to estimate channel parameters and then relies on the maximum-likelihood concept to detect molecular signals. However, the essence of these schemes is to assume an exact channel model with correct channel parameter estimation, in order to compute the CSI and their likelihood densities for detection. In other words, if the estimations of these parameters are biased, an erroneous inference of the digital information is inevitable. Moreover, we consider some adverse cases where the channel is complex and even the expression of the CIR is unavailable (e.g., an absorbing channel or channels with micro-fluid and obstacles). It is undeniable that one can resort to the stochastic methods e.g., Markov Chain Monte Carlo (MCMC) \cite{gilks1995markov} to estimate the CSI and then compute the likelihood densities. They will expense at the huge expenditures of the computation and storage resources, as well as suffering from detection errors caused by potentially erroneous estimations of the CSI.

Alternatively, the other group of detection method is inspired by nature. We can list the enzyme equalizer \cite{6712164, 7925961} and the stochastic resonance (SR) filter \cite{8255066}, which are all non-coherent communication schemes. In our previous papers \cite{8255066, 7353215, 7435284}, we proposed three metrics for non-coherent detectors, whereby the intrinsic properties of molecular signals are extracted from the received data, which then are linearly combined for information inference. However, the linear combinations may cause a loss of intrinsic and characteristic features, and thereby limit their detection performances. Also, there is still a lack of understanding on the theoretical performance bound of the proposed high-dimensional non-coherent detector. These two open challenges constitute the motivation to extend our previous research.

\subsection{Contributions}
In this work, we suggest a novel non-coherent detection scheme based on a newly designed high-dimensional metric, aiming to build a blind detection paradigm for MCvD. To sum up, the main contributions of this paper are listed as follows:

(1) A high-dimensional metric is designed via an exploration of the inherently transient features of the molecular signals. Compared with the coherent schemes that focus on the total CSI, the metric concentrates only on the obvious differences invoked by 1-bit and 0-bit, therefore making it easier to estimate its likelihood density in the blind detection. Also, as opposed to our previously linear non-coherent works in \cite{8255066, 7353215, 7435284}, the high-dimensional metric can improve the detection performance, as it has a greater signal-to-noise ratio (SNR) by constructing a spikier likelihood function.

(2) From the theoretical perspective, we compute the theoretical bit error rate (BER) for any designed high-dimensional non-coherent scheme premised on Bayesian inference. Also, we prove that, as the high-dimensional non-coherent metrics are constructed by the same samples, the metric with higher dimensions has a lower BER. This further indicates that our proposed high-dimensional non-coherent scheme outperforms the previously linear ones, and that its lower-bound of BER converges to that of the coherent MAP.

(3) For the blind detection, we suggest the Parzen technique with a probabilistic neural network (Parzen-PNN) to approximate the likelihood density, and further detect the information bits. Using a Gaussian Parzen window, the Parzen-PNN is capable of approaching the theoretical BER. Moreover, compared with other machine learning schemes for classification, Parzen-PNN takes the advantages of fast convergence and the parallel implementation \cite{li2011novel,duda2012pattern,specht1990probabilistic}. Therefore, it is capable of ensuring the real-time communication and also meeting the requirement of the limited energy-expenditure for the nano-receiver.

(4) We evaluate the detection performance of our proposed high-dimensional non-coherent scheme via simulations. The results demonstrate that the performance of signal detection is improved about $10$dB in terms of the BER, as opposed to our previous algorithm, and the MAP with the unknown CSI. This suggests that the proposed scheme presents great promise to its MCvD detection. Moreover, attributed to its blind detection, our proposed scheme also can be modified and thereby has a potential impact on other signal detection applications.

The rest of this article is structured as follows. In Section II, the system structure of the MCvD is specified, along with a short description of the state-of-the-art MAP and our previous linear non-coherent schemes. In Section III, we elaborate our designed high-dimensional non-coherent scheme, compute its theoretical BER, and prove that it has a better communication performance than the low-dimensional non-coherent schemes. In Section IV, we introduce the Parzen-PNN with the aim of the realization of our scheme. Numerical simulations are provided in Section V. Finally, we conclude this study in Section VI along with discussions of application areas and future impact.

\section{System Models}
\subsection{Molecular Communications}
Similar to an EMW-based communication system, a generalized model for MCvD is illustrated in Fig. 1, consisting of a nano-transmitter, a propagation channel, and a nano-receiver.

\subsubsection{Nano-transmitter}
The nano-transmitter may be either a single cell/organism in a biological system, or an artificially designed hardware, and aims to modulate the ON/OFF key (OOK) information bit $\alpha_k=j\in\{0,1\}~(k=0,\cdots,+\infty)$ via molecular amplitude (i.e., the concentration or the number of a specific type of molecules) or phase (i.e., interval) \cite{PURVIS2013945}. Here, we consider the amplitude modulation (AM) from \cite{7553035}, and the modulated signal, denoted as $s(t)$ can be thereby expressed as:
\begin{equation}
s(t)=Q\cdot\sum_{k=0}^\infty\alpha_k\cdot\delta(t-kT_b),
\end{equation}
where $T_b$ is the bit interval, and $Q$ represents the emitted concentration of the molecules. Here, $\delta(\cdot)$ accounts for the Dirac function that is adopted to describe the pulse shape.

\begin{figure*}[!t]
\centering
\includegraphics[width=7in]{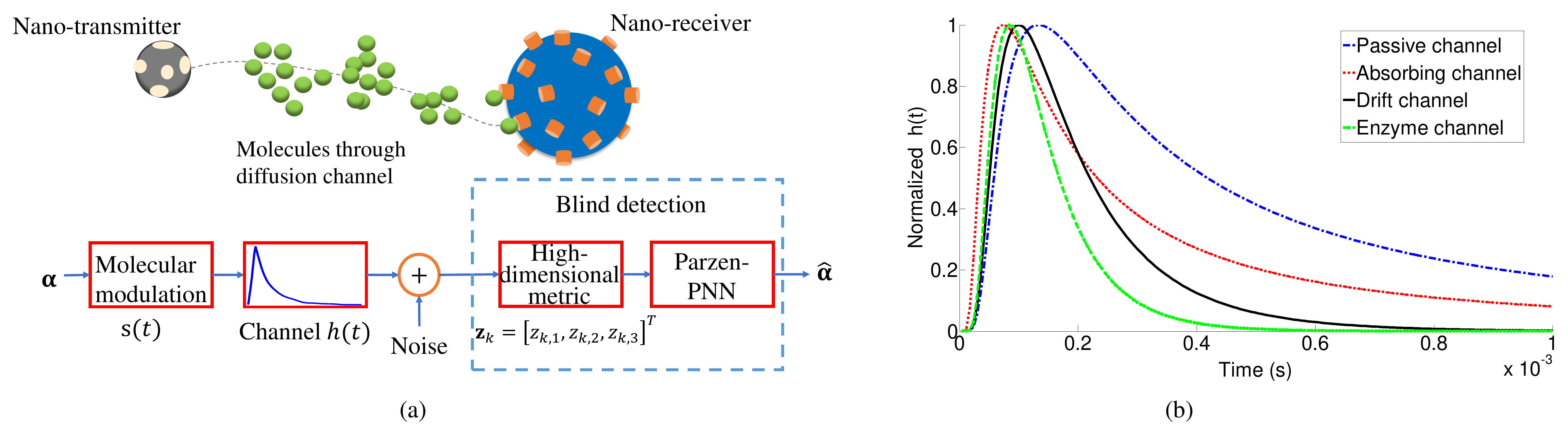}
\caption{Illustration of MCvD system. (a) gives the schematic flow where the binary information $\bm{\alpha}=[\alpha_0,\cdots,\alpha_k]^T$ are modulated by the Nano-transmitter via molecules, which will then propagate through the diffusive channel $h(t)$, and received by the nano-receiver. After sampling the received signal, the receiver will construct the high-dimensional metric and then rely on the Parzen-PNN algorithm to detect the information $\hat{\bm{\alpha}}=[\hat{\alpha}_0,\cdots,\hat{\alpha}_k]^T$. (b) shows the different cases of diffusive channels and their corresponding CIRs}
\end{figure*}

\subsubsection{Propagation channel}
Typically, a diffusive channel model and its corresponding CIR (denoted as $h(t)$) remains unknown due to the infeasible computations of the complex channel properties (e.g., a mixture with absorbing and obstacles, presence of advection forces, fluctuating channel parameters...etc.). Hence, we assume an unknown CIR for our high-dimensional non-coherent signal detection, by randomly selecting one of the two widely-used diffusion models in Eq. (2) for performance analysis and comparisons with the state-of-the-art MAP. The two models (including the passive and the absorbing channels) and their CIRs are given as follows \cite{6807659, 6168841, 6712164, 7841566}:
\begin{equation}
h(t)=
\begin{cases}
{\huge\frac{V}{(4\pi Dt)^{3/2}}}\exp\left(-\frac{(r-vt)^2}{4Dt}-\beta t\right) & \text{passive}\\[2mm]
{\huge\frac{R}{R+r}\frac{r}{\sqrt{4\pi Dt^3}}}\exp\left(-\frac{(r-vt)^2}{4Dt}-\beta t\right) & \text{absorbing}\\
\end{cases}
\end{equation}
where $V$ gives the volume of a spherical nano-receiver with its radius $R$, $r$ is the transmitted distance, $D$ represents the diffusion coefficient, $v$ is the velocity of the drift, and $\beta$ is the degradation exponent of the enzyme interactions.

As we assign various parameters for Eq. (2), we can see the different CIRs in Fig. 1(b). It is noteworthy that although the exact shapes of the CIRs depends heavily on the model expressions and their model-related parameters, there are still transient features in common, e.g., an obvious rising edge for the appearance of an 1-bit, the inflexions, and the concentration difference between two adjacent bits, which may give hints for the design of a blind detection paradigm in the receiver.

\subsubsection{Nano-receiver}
At the receiver end, the received signal is expressed as:
\begin{equation}
y(t)=h(t)\otimes s(t)+\epsilon(t).
\end{equation}
Here, Eq. (3) accounts for a linear system model \cite{6708551, 7435284}, whereby the CIR convolutes the input $s(t)$ (via convolution sign $\otimes$) to create a linear combination of recorded data. $\epsilon(t)$ is the additive noise induced by the imperfect counting process or other environmental disturbances, following an independent and identically distributed (i.i.d.) distributions\footnote{The additive noise $\epsilon(t)$ may follow the i.i.d. Poisson distribution \cite{7331288}, and may be approximated as the Gaussian distribution as the number of molecules grows larger \cite{5452953}. In this paper, it is noteworthy that we use only the condition of the i.i.d distribution for analysis.} \cite{7331288,5452953}.

After receiving $y(t)$, the nano-receiver will sample the concentration of signaling molecules under the sample rate $T_s$ (subject to Nyquist theorem \cite{6708551, 7435284} with $MT_s=T_b$). Without losing the generality, we assume the synchronization has been accurately accomplished via the techniques in \cite{6626319}, and hence the discrete signal within time interval $iT_s\in[kT_b,(k+1)T_b)$ readouts:
\begin{equation}
y_i=\alpha_kh_{i-kM}+\sum_{l=0}^{k-1}\alpha_l\cdot h_{i-lM}+\epsilon_i,
\end{equation}
with $y_i=y(iT_s)$, $h_i=h(iT_s)$, and $\epsilon_i=\epsilon(iT_s)$.

We can observe from Eq. (4) that the ISI is induced due to the long-tail nature of $h(t)$, and will deteriorate the current detection of $\alpha_k$. This will be even worse as we do not know the explicit expression of CIR $h(t)$, and the combined effect from the ambient noise. As such, the purpose of this paper is to detect the current informative bit $\alpha_k$ from the received signal $\mathbf{y}_{0:(k+1)M-1}=[y_0,...,y_{(k+1)M-1}]^T$ in a blind detection mode.

\subsection{Two State-of-the-Art Methods for Comparison}
\subsubsection{Coherent MAP}
In general, an MAP detector aims at maximizing \emph{a posteriori} of the unknown information bits conditioned on the received samples, i.e.,
\begin{equation}
\begin{aligned}
\hat{\bm{\alpha}}_\text{MAP}&
=\operatorname*{argmax}\limits_{\bm{\alpha}}p(\bm{\alpha}|\mathbf{y})\\
&=\operatorname*{argmax}\limits_{\bm{\alpha}}\prod_{l=0}^kp(\alpha_l|\bm{\alpha}_{0:l-1})
\prod_{l=0}^kp(\mathbf{y}_{Ml:M(l+1)-1}|\bm{\alpha}_{0:l}).
\end{aligned}
\end{equation}
However, for blind detection, in the above coherent MAP, the accurate estimation of the CIR will be indispensable in computing its likelihood density and its transitional probability for the Viterbi tracing. This will be increasingly difficult as one do not know the expression of $h(t)$, and serious computation and storage resources will be spent in estimating the CIR \cite{7435284}.

\subsubsection{Our previous non-coherent method}
In essence, the previously designed non-coherent method aims at amplifying the SNR via a linear combination of three metrics (i.e., $c_{k,1}$, $c_{k,2}$ and $c_{k,3}$) that explore the transient features of molecular signals, i.e. \cite{8255066},
\begin{equation}
c_k^{\text{linear}}=c_{k,1}+c_{k,2}+c_{k,3}.
\end{equation}
In Eq. (6), $c_{k,1}$ is referred as the local geometry shape that characterizes the maximum inflexion induced by 1-bit, $c_{k,2}$ describes the inflexion caused by the new arrival of 1-bit, while $c_{k,3}$ gives the concentration difference between successive bits. However, two drawbacks still remain. Firstly, $c_{k,1}$ performs poorly as the ISI increases, due to the fluctuating inflexions in the presence of a strong ISI. Secondly and more importantly, the effect of the SNR amplification is not promising, due to the limitation of the 1-dimensional metric (which will be analyzed in Section III. D).

\section{High-dimensional Metric based non-coherent detection}
In this section, we design a high-dimensional non-coherent scheme aiming to build a blind detection scheme for MCvD. We firstly construct the high-dimensional metric by re-designing and re-combining the metrics that can explore the molecular transient features better. Then, we compute the high-dimensional decision surface via Bayesian inference, based on which the detection can be pursued.

\begin{figure*}[!t]
\centering
\includegraphics[width=6in]{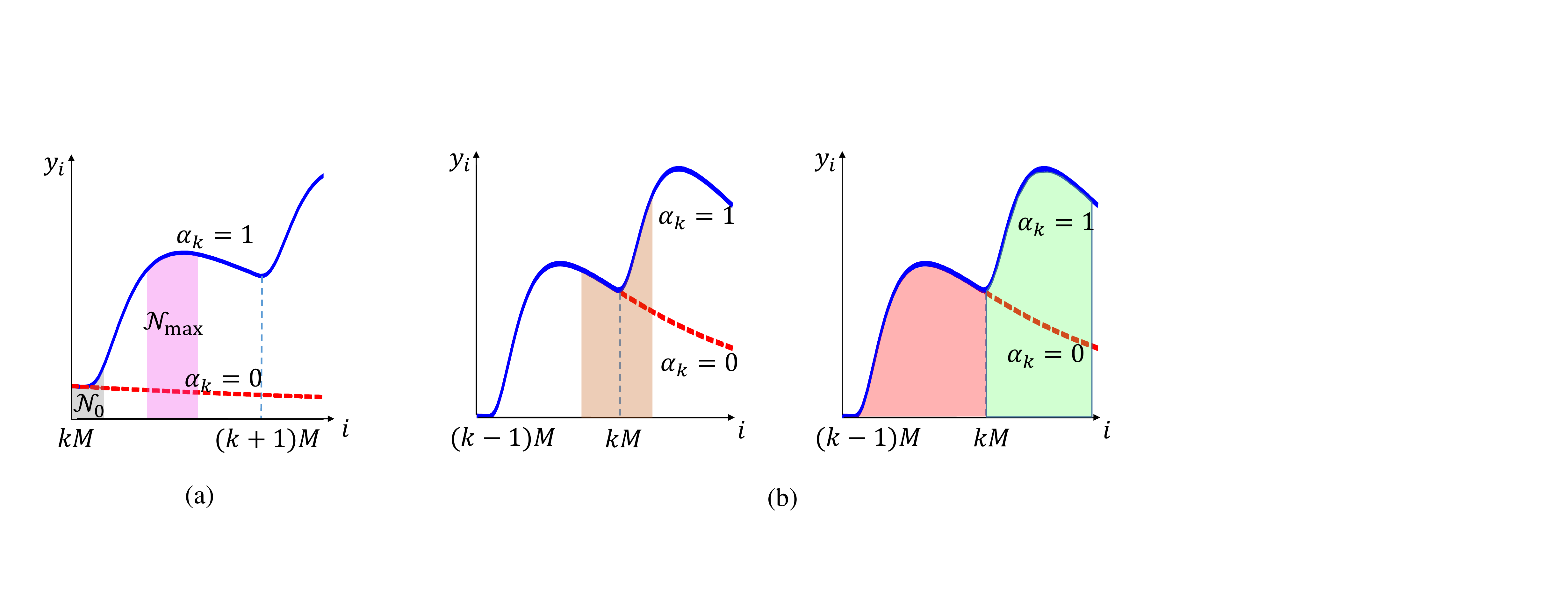}
\caption{Illustration of sub-metrics. (a) gives the local rising edge metric $z_{k,1}$, whilst (b) shows the successive properties that contains the inflexion metric $z_{k,2}$ and the concentration difference $z_{k,3}$}
\end{figure*}

\subsection{Construction of High-dimensional Metric }
In comparison with the linear metric in Eq. (6), the $d$-dimensional metric denoted as $\mathbf{z}_k$ is a vector composed of $d$ sub-metrics (e.g., the rising edge, and the successive properties between adjacent symbols) that can describe the transient features of the signal. Here, we express $\mathbf{z}_k$ via a $d\times (k+1)M$ transformation matrix $\mathbf{A}_{d\times (k+1)M}$, i.e.,
\begin{equation}
\mathbf{z}_k=\mathbf{A}_{d\times (k+1)M}\cdot\mathbf{y}_{0:(k+1)M-1},
\end{equation}
where each row of $\mathbf{z}_k$ denotes one sub-metric.

In the context of the MCvD, $d=3$ sub-metrics are designed by respectively exploring \textbf{1)} the \emph{local rising-edge} in each symbol, denoted as $z_{k,1}$, \textbf{2)} the two \emph{successive properties} between two adjacent symbols, denoted as $z_{k,2}$ and $z_{k,3}$, i.e. ,
\begin{equation}
\mathbf{z}_k=[z_{k,1},z_{k,2},z_{k,3}]^T.
\end{equation}

\subsubsection{Local rising-edge} Taking the $k$th interval with $M=T_b/T_s$ samples as an example, in the case of $\alpha_k=1$, the output $y_i$ will have a distinct rising edge. As is shown in Fig. 2(a), this rising edge can be expressed by the difference of its maximum (computed by averaging its neighbourhood $\mathcal{N}_{\text{max}}$) from the beginning $\mathcal{N}_{\text{0}}$. In practice, by specifying the widths of $\mathcal{N}_{\text{max}}$ and $\mathcal{N}_{\text{0}}$ to be $|\mathcal{N}_{\text{max}}|=|\mathcal{N}_{\text{0}}|=M/4$, we define the metric of the local rising-edge as:
\begin{equation}
z_{k,1}\triangleq\frac{1}{|\mathcal{N}_{\text{max}}|}\sum_{i\in\mathcal{N}_{\text{max}}}y_i
-\frac{1}{|\mathcal{N}_{\text{0}}|}\sum_{i\in\mathcal{N}_{\text{0}}}y_i.
\end{equation}

It is easily noted that in the case of $\alpha_k=1$, $z_{k,1}$ will be larger than $0$, otherwise, it will be smaller that $0$ when $\alpha_k=0$. Also, compared with the previously designed geometry shape i.e., $c_{k,1}$ in Eq. (6), the rising edge (i.e., $z_{k,1}$) will not disappear with an increasing intensity of the ISI, thereby capable of characterising the transient feature of the molecular signals in different conditions of the ISI. In these views, $z_{k,1}$ can be employed as a high-performance metric to distinguish whether there are new arrivals of molecules induced by 1-bit at the nano-receiver at current time interval.

\subsubsection{Successive properties}
When it comes to two successive slots $k-1$ and $k$, the transient features will be quite different with respect to two cases (i.e., $\alpha_{k}=1$ and $\alpha_k=0$). Here, we adopt our previously designed sub-metrics (i.e., the minimum inflexion $c_{k,2}$ and  the concentration-difference $c_{k,3}$, as is shown in Fig. 2(b)), i.e.,
\begin{equation}
z_{k,2}\triangleq c_{k,2},
\end{equation}
\begin{equation}
z_{k,3}\triangleq c_{k,3},
\end{equation}
where the expressions of $c_{k,2}$ and $c_{k,3}$ are referred from \cite{8255066}.

\subsection{Distribution of High-dimensional Metric}
Given that the noisy samples $y_i$ of various discrete time follow i.i.d distribution, the sub-metrics from Eqs. (9)-(11) can be regarded as Gaussian random variables (RVs), due to the central limit theorem (CLT) with a large sample rate ($M\geq50$). Hence, the $d$-dimensional $\mathbf{z}_k$ follows the multivariate normal distribution, i.e.,
\begin{equation}
\mathbf{z}_k\sim
\begin{cases}
\mathcal{N}(\bm{\mu}_1,\bm{\Sigma}), & \alpha_k=1,\\[2mm]
\mathcal{N}(\bm{\mu}_0,\bm{\Sigma}), & \alpha_k=0.\\
\end{cases}
\end{equation}
where $\bm{\mu}_j=\mathbb{E}(\mathbf{z}_k|\alpha_k=j)$ represents the mean of $\mathbf{z}_k$. $\bm{\Sigma}$ gives the covariance matrix.

Then, according to Eq. (12), the likelihood probability density function (PDF) of the $d$-dimensional metric $\mathbf{z}_k$ conditioned on the different information bits (i.e., $\alpha_k=1$ or $\alpha_k=0$) are:
\begin{equation}
\!\!\varphi(\mathbf{z}_k|\alpha_k)=\frac{1}{\sqrt{(2\pi)^d|\bm{\Sigma}|}}\exp\left(-\frac{1}{2}(\mathbf{z}_k-\bm{\mu}_j)^T\bm{\Sigma}^{-1}(\mathbf{z}_k-\bm{\mu}_j)\right).\!\!
\end{equation}
Here, it is noteworthy that the likelihood PDF in Eq. (13) also accounts for any designed $d$-dimensional metric $\mathbf{z}_k$ in Eq. (7), with its mean $\bm{\mu}_j$ and variance $\bm{\Sigma}$.

\subsection{High-dimensional Decision Surface}
After specifying the form of the $d$-dimensional metric and its distributions, we then study the detection process. By adopting the Bayesian rule, the detection process is equivalent to compute and compare the posterior PDFs of $\mathbf{z}_k$ from different cases of $\alpha_k=0$ and $\alpha_k=1$, i.e.,
\begin{equation}
\frac{p(\alpha_k=1|\mathbf{z}_k)}{p(\alpha_k=1|\mathbf{z}_k)}\mathop{\gtreqless}\limits^{\hat{\alpha}_k=1}\limits_{\hat{\alpha}_k=0}1,
\end{equation}
where $p(\alpha_k|\mathbf{z}_k)$ represents the posterior PDF.

Then, as we take $p(\alpha_k|\mathbf{z}_k)=\varphi(\mathbf{z}_k|\alpha_k)\cdot \text{Pr}\{\alpha_k\}$ into Eq. (14), we can specify the decision surface in the form of logarithm difference, i.e.,
\begin{equation}
\ln\varphi(\mathbf{z}_k|\alpha_k=1)-\ln\varphi(\mathbf{z}_k|\alpha_k=0)=0,
\end{equation}
with an assumption of an identically prior probability as $\text{Pr}\{\alpha_k=0\}=\text{Pr}\{\alpha_k=1\}=0.5$. Taken Eq. (13) into the left-hand of Eq. (15), the decision surface is computed as:
\begin{equation}
\mathcal{G}(\mathbf{z}_k)=0,
\end{equation}
with
\begin{equation}
\mathcal{G}(\mathbf{z}_k)=(\bm{\mu}_1^T-\bm{\mu}_0^T)\bm{\Sigma}^{-1}\mathbf{z}_k
-\frac{1}{2}(\bm{\mu}_1^T\bm{\Sigma}^{-1}\bm{\mu}_1-\bm{\mu}_0^T\bm{\Sigma}^{-1}\bm{\mu}_0).
\end{equation}
As such, the detection can be solved by analyzing whether $\mathbf{z}_k$ is above the decision surface in Eqs. (16)-(17).

Also, it is noteworthy that \textbf{1)} this decision surface holds for the general form of any designed $d$-dimensional metric, suggesting that a general expression of the theoretical BER may exist (which will be computed in the next part), and \textbf{2)} for the blind detection, parameters in Eq. (17) e.g., $\bm{\mu}_j$ and $\bm{\Sigma}$ are model-related and unknown, which will be discussed in Section IV.

\subsection{Theoretical Performance Analysis}
As we derive the general form of decision surface in Eqs. (16)-(17), and the general likelihood PDF in Eq. (13) for any $d$-dimensional metric $\mathbf{z}_k$, we here compute the theoretical BER, and analyze the theoretical performance among different non-coherent schemes.

\subsubsection{Computation of BER}
The theoretical BER of the $d$-dimensional $\mathbf{z}_k$ can be computed in terms of the two identical error probabilities, i.e.,
\begin{equation}
\begin{aligned}
\!P_\text{e}(\mathbf{z}_k)&=\underset{\mathcal{G}(\mathbf{z})\geqslant0}{\overset{d}{\overbrace{\int\cdots\int}}}\varphi(\mathbf{z}|\alpha_k=0)d\mathbf{z}
+\underset{\mathcal{G}(\mathbf{z})<0}{\overset{d}{\overbrace{\int\cdots\int}}}\varphi(\mathbf{z}|\alpha_k=1)d\mathbf{z}\!\\
&=\!2\underset{\mathcal{G}(\mathbf{z})\geqslant0}{\overset{d}{\overbrace{\int\cdots\int}}}\frac{1} {\sqrt{(2\pi)^d|\bm{\Sigma}|}}\exp\left(-\frac{1}{2}\mathbf{z}^T\bm{\Sigma}^{-1}\mathbf{z}\right)dz_1\cdots z_d\!\\
&\overset{(\text{i})}{=}\!\!2\!\underset{\mathcal{G}(\bm{\Gamma}\cdot\mathbf{x})\geqslant0}{\overset{d}{\overbrace{\int\cdots\int}}}\frac{|\mathbf{J}|}{\sqrt{(2\pi)^d|\bm{\Sigma}|}}\exp\left(-\frac{1}{2}\mathbf{x}^T\bm{\Lambda}^{-1}\mathbf{x}\right)dx_1\cdots x_d\!\!\!\\
&\overset{(\text{ii})}{=}\!2\underset{\mathcal{G}(\bm{\Gamma}\cdot\bm{\Lambda}^{0.5}\cdot\mathbf{u})\geqslant0}{\overset{d}{\overbrace{\int\cdots\int}}}\frac{1}{\sqrt{(2\pi)^d}}\exp\left(-\frac{1}{2}\mathbf{u}^T\mathbf{u}\right)du_1\cdots u_d\!\\
&\overset{(\text{iii})}{=}2\cdot\Phi\left(-\frac{1}{2}\sqrt{(\bm{\mu}_1-\bm{\mu}_0)^T\bm{\Sigma}^{-1}(\bm{\mu}_1-\bm{\mu}_0)}\right).
\end{aligned}
\end{equation}
The explanation of Eq. (18) is given as follows.

(i) As we observe that $\mathbf{z}^T\cdot\bm{\Sigma}\cdot\mathbf{z}$ holds for the quadratic form with $\bm{\Sigma}$ a $d$-dimensional symmetric matrix, we have $\mathbf{z}=\bm{\Gamma}\cdot\mathbf{x}$, and $\mathbf{z}^T\cdot\bm{\Sigma}\cdot\mathbf{z}=\mathbf{x}^T\cdot\bm{\Lambda}\cdot\mathbf{x}$, where $\bm{\Lambda}=\text{diag}(\lambda_1,\cdots,\lambda_d)$ is composed of $d$ different eigenvalues of $\bm{\Sigma}$, and $\bm{\Gamma}$ is composed of the corresponding $d$ normalized eigenvectors, such that $\bm{\Sigma}=\bm{\Gamma}\cdot\bm{\Lambda}\cdot\bm{\Gamma}^{-1}$. Hence, by replacing $\mathbf{z}$ with $\mathbf{x}$, the equation of (i) is achieved with the the help of the Jacobian determinant, as
\begin{equation}
\left|\mathbf{J}\right|=\begin{vmatrix}\frac{\partial z_1}{\partial x_1} & \cdots&\frac{\partial z_1}{\partial x_d}\\ \vdots & \ddots & \vdots\\ \frac{\partial z_d}{\partial x_1} & \cdots &\frac{\partial z_d}{\partial x_d}\\\end{vmatrix}
=|\bm{\Gamma}|=1.
\end{equation}

(ii) We here use $\mathbf{u}=\bm{\Lambda}^{-0.5}\mathbf{x}$ to replace $\mathbf{x}$. Then, by noticing that $|\bm{\Sigma}|=|\bm{\Lambda}|=\prod_{i=1}^d\lambda_i$, we can obtain the equation of (ii).

(iii) Before we explain the establishment of (iii), we firstly prove the equality as follows:
\begin{equation}
\int\limits_{-\infty}^{+\infty}\frac{1}{\sqrt{2\pi}}\exp\left(-\frac{z^2}{2}\right)\cdot\Phi\left( az+b\right)dz
\equiv\Phi\left(\frac{b}{\sqrt{1+a^2}}\right),
\end{equation}
where
\begin{equation}
\Phi(x)\triangleq\int_{-\infty}^x\frac{1}{\sqrt{2\pi}}\exp\left(-\frac{z^2}{2}\right)dz.
\end{equation}
This is because
\begin{equation}
\begin{aligned}
&\int\limits_{-\infty}^{+\infty}\frac{1}{\sqrt{2\pi}}\exp\left(-\frac{z^2}{2}\right)\cdot\Phi\left( az+b\right)dz\\
=&\int_{-\infty}^b\frac{\partial\left(\int_{-\infty}^{+\infty}\frac{1}{\sqrt{2\pi}}\exp\left(-\frac{z^2}{2}\right)\cdot\Phi\left( az+x\right)dz\right)}{\partial x}dx\\
=&\int_{-\infty}^b\frac{1}{\sqrt{2\pi(1+a^2)}}\exp\left(-\frac{x^2}{2(1+a^2)}\right)dx\\
=&\Phi\left(\frac{b}{\sqrt{1+a^2}}\right).
\end{aligned}
\end{equation}
Then, by taking Eq. (20) back to (ii), we can prove the (iii) via
\begin{equation}
\begin{aligned}
&\!2\cdot\underset{\mathcal{G}(\bm{\Gamma}\cdot\bm{\Lambda}^{0.5}\cdot\mathbf{u})\geqslant0}{\overset{d}{\overbrace{\int\cdots\int}}}\frac{1}{\sqrt{(2\pi)^d}}\exp\left(-\frac{1}{2}\mathbf{u}^T\mathbf{u}\right)du_1\cdots u_d\!\\
=&2\cdot\int\limits_{-\infty}^{+\infty}\frac{1}{\sqrt{2\pi}}\exp\left(-\frac{u_1^2}{2}\right)
\cdots\int\limits_{-\infty}^{+\infty}\frac{1}{\sqrt{2\pi}}\exp\left(-\frac{u_{d-1}^2}{2}\right)\\
&\!\cdot\Phi\left(\frac{\xi_{d-1}}{\xi_d}u_{d-1}-\frac{\xi_d}{2}+\sum_{i=1}^{d-2}\left(\frac{\xi_i}{\xi_d}u_i-\frac{\xi_i^2}{2\xi_d}\right)\right)du_1\cdots du_{d-1}\!\\
=&2\cdot\Phi\left(-\frac{1}{2}\sqrt{\bm{\xi}^T\cdot\bm{\xi}}\right),
\end{aligned}
\end{equation}
where $\bm{\xi}=[\xi_1,\ldots,\xi_d]^T=\bm{\Lambda}^{-0.5}\bm{\Gamma}^{-1}\cdot(\bm{\mu}_1-\bm{\mu}_0)$.

Hence, in Eq. (18), we give the general form of BER for any designed $d$-dimensional metric $\mathbf{z}_k$, as one can take the metric-related $\bm{\mu}_j$ and $\bm{\Sigma}$ into Eq. (18). For instance, in our analysis of the MCvD, we can derive the theoretical BER of our $d=3$ metric as well as the one of our previously proposed linear ($d=1$) metric in \cite{8255066}, and their BER performances can be compared from the theoretical perspective.

\subsubsection{Comparison among non-coherent schemes}
We compare the BERs between the non-coherent schemes with metrics as a designed $d$-dimensional $\mathbf{z}_k$, and a $(d_1\leq d)$-dimensional metric, i.e., $\mathbf{A}\cdot\mathbf{z}_k$. Here, we assume $rank(\mathbf{A})=d_1$, otherwise its dimension can be reduced such that the determinant of its covariance matrix is positive, i.e., $|\bm{\Sigma}(\mathbf{A}\cdot\mathbf{z}_k)|>0$. We give the comparison result as:
\begin{equation}
P_{\text{e}}(\mathbf{z}_k)\leq P_{\text{e}}(\mathbf{A}\cdot\mathbf{z}_k),
\end{equation}
where the equality holds as $d_1=d$.

The proof of the Eq. (24) is given as follows. Given Eq. (18), and the monotonically increasing property of $\Phi(x)$ with respect to $x$, the comparison between $P_{\text{e}}(\mathbf{z}_k)$ and $ P_{\text{e}}(\mathbf{A}\cdot\mathbf{z}_k)$ can be converted as analyzing the quadratic values of $\bm{\mu}^T(\bm{\Sigma}^{-1}-\mathbf{A}^T(\mathbf{A}\bm{\Sigma}\mathbf{A}^T)^{-1}\mathbf{A})\bm{\mu}$ with $\bm{\mu}=\bm{\mu}_1-\bm{\mu}_0$ and $\bm{\Sigma}(\mathbf{A}\cdot\mathbf{z}_k)=\mathbf{A}\bm{\Sigma}\mathbf{A}^T$. In this view, Eq. (24) is equivalent with $\bm{\Sigma}^{-1}-\mathbf{A}^T(\mathbf{A}\bm{\Sigma}\mathbf{A}^T)^{-1}\mathbf{A}$ being positive semidefinite.

(i) In the case of $rank(\mathbf{A})=d_1=d$, we have
\begin{equation}
\bm{\Sigma}^{-1}-\mathbf{A}^T(\mathbf{A}\bm{\Sigma}\mathbf{A}^T)^{-1}\mathbf{A}=\mathbf{0}
\end{equation}
which proves the equality of the Ineq. (24)

(ii) For $d_1<d$, we consider only $d_1=d-1$, and other cases $d_1<d-1$ can be proved successively by replacing $d$ as $d-1$. We firstly divide $\mathbf{A}=\mathbf{A}_1\cdot\bm{\Gamma}^{-1}$ (where $\bm{\Gamma}$ is the matrix of $d$ eigenvectors of $\bm{\Sigma}$, and therefore $rank(\bm{\Gamma})=d$), with $\mathbf{A}_1=\mathbf{A}\cdot\bm{\Gamma}$. Then, the metric can be converted as $\mathbf{A}\cdot\mathbf{z}_k=\mathbf{A}_1\cdot\bm{\Gamma}^{-1}\mathbf{z}_k$ whereby the covariance matrix of $\bm{\Gamma}^{-1}\cdot\mathbf{z}_k$ is $\bm{\Sigma}(\bm{\Gamma}^{-1}\cdot\mathbf{z}_k)=\bm{\Lambda}$. Also, note that $\mathbf{A}_1=\bm{\Pi}\cdot\mathbf{A}_2$, with $\bm{\Pi}$ the multiplications of elementary row transformations ($rank(\bm{\Pi})=d-1$), and $\mathbf{A}_2=[\mathbf{I}_{d_1\times d_1}~\mathbf{a}]$ where $\mathbf{a}=[a_1,\cdots a_{d-1}]^T$. Hence, according to (i), with $P_{\text{e}}(\bm{\Gamma}^{-1}\mathbf{z}_k)=P_{\text{e}}(\mathbf{z}_k)$, and $P_{\text{e}}(\mathbf{A}\cdot\mathbf{z}_k)=P_{\text{e}}(\bm{\Pi}\cdot\mathbf{A}_2\cdot\bm{\Gamma}^{-1}\mathbf{z}_k)=P_{\text{e}}(\mathbf{A}_2\cdot\bm{\Gamma}^{-1}\mathbf{z}_k)$, we only need to prove that $\bm{\Lambda}^{-1}-\mathbf{A}_2^T(\mathbf{A}_2\bm{\Lambda}\mathbf{A}_2^T)^{-1}\mathbf{A}_2$ is a positive semidefinite matrix. This can be illustrated as it has only non-negative eigenvalues:
\begin{equation}
\begin{aligned}
&\bm{\Lambda}^{-1}-\mathbf{A}_2^T\left(\mathbf{A}_2\bm{\Lambda}\mathbf{A}_2^T\right)^{-1}\mathbf{A}_2\\
=&\bm{\Lambda}^{-1}-\mathbf{A}_2^T\left(diag(\lambda_1,\ldots,\lambda_{d-1})^{-1}-\zeta\cdot\bm{\varrho}\bm{\varrho}^T\right)\mathbf{A}_2\\
=&\zeta\cdot
\begin{bmatrix}
\bm{\varrho}\bm{\varrho}^T & -\bm{\varrho}\\
-\bm{\varrho}^T & 1\\
\end{bmatrix}\\
=&\zeta\cdot\bm{\Upsilon}\cdot diag\left(tr\left(\begin{bmatrix}
\bm{\varrho}\bm{\varrho}^T & -\bm{\varrho}\\
-\bm{\varrho}^T & 1\\
\end{bmatrix}\right),0,\cdots0\right)\cdot\bm{\Upsilon}^{-1}\\
=&\bm{\Upsilon}\cdot diag\left(\zeta\cdot tr\left(1+\bm{\varrho}^T\bm{\varrho}\right),0,\cdots0\right)\cdot\bm{\Upsilon}^{-1},
\end{aligned}
\end{equation}
with $\bm{\varrho}=[a_1\sqrt{\lambda_d}/\lambda_1,\ldots,a_{d-1}\sqrt{\lambda_d}/\lambda_{d-1}]^T$, $\zeta=1/(1+\lambda_d\sum_{i=1}^{d-1}a_i^2/\lambda_i)$, and $\bm{\Upsilon}$ the eigenvector matrix, which therefore proves the Ineq. (24).

The meaning of the Eq. (24) is explained as follows. First, it demonstrates that via using the same samples, the metric with higher dimension has the lower BER, i.e.,
\begin{equation}
\begin{aligned}
P_{\text{e}}(&\mathbf{A}_{d\times (k+1)M}\cdot\mathbf{y}_{0:(k+1)M-1})\\
&<P_{\text{e}}(\mathbf{A}_{(d_1<d)\times d}\mathbf{A}_{d\times (k+1)M}\cdot\mathbf{y}_{0:(k+1)M-1}).
\end{aligned}
\end{equation}
This further indicates that our designed $3$-dimensional metric $\mathbf{z}_k=[z_{k,1},z_{k,2},z_{k,3}]^T$ outperforms the previous linear one in \cite{8255066} as $z_{k,1}+z_{k,2}+z_{k,3}$, i.e.,
\begin{equation}
P_{\text{e}}(\mathbf{z}_k)<P_{\text{e}}(z_{k,1}+z_{k,2}+z_{k,3}).
\end{equation}

Secondly, the Eq. (24) indicates that the lower-bound of any designed metric $\mathbf{z}_k$ is the BER of the MAP under the known of the CIR, i.e.,
\begin{equation}
P_{\text{e}}(\mathbf{z}_k)\geq P_{\text{e}}(\mathbf{y}_{0:K-1})\geq P_{\text{e}}(\text{MAP}),
\end{equation}
as the Viterbi tracing algorithm can be pursued to reduce the BER by counteracting the effect of the ISI.

However, we should also notice two aspects. For one thing, the coherent MAP can only reach its theoretical BER if the CIR $h(t)$ is known. Otherwise, its performance depends directly on how accurate the estimation of the CIR is, which is difficult and will consume most of its computational resources \cite{7435284}. Secondly, there is a resource-performance trade-off as we select and design the proper metric for the blind detection process. The performance of $d$-dimensional non-coherent scheme betters as the $d$ grows larger, yet at the expense of both computational and storage complexities with respect to $d$, especially when we need to estimate or approximate its $d$-dimensional likelihood density.

\section{Parzen-PNN realization}
It is noteworthy that the decision surface in Eq. (14) is only computable in the cases of a known CIR $h(t)$, since $\bm{\mu}_j$ and $\bm{\Sigma}$ from the likelihood PDF in Eq. (12) depend on the expression of $h(t)$. Otherwise, in most complex MCvD scenarios where $h(t)$ is unavailable, the detection process will be malfunctioning as one cannot derive the likelihood distribution. In this view, it is demanding to resort to an alternative method that is capable of approximating (or estimating) the likelihood PDF in Eq. (13). And here come the Parzen window technique and its PNN based implementation.

\subsection{Parzen window technique}
In essence, the Parzen-windowing technique \cite{parzen1962estimation} approximates (or estimates) the probability by defining a window (given the window size) and a function
on this window (i.e. referred as the window function). Given a $d$-dimensional RV $\mathbf{z}$ with unknown distribution, Parzen-windowing estimates its corresponding PDF $p(\mathbf{z})$ by sampling its observations within the window function $\Pi(\mathbf{z})$, i.e.\cite{parzen1962estimation, li2011novel},
\begin{equation}
p(\mathbf{z})\simeq\frac{1}{N}\sum_{n=1}^N\frac{1}{\varsigma^d}\cdot\Pi\left(\frac{\mathbf{z}-\mathbf{z}^{(n)}}{\varsigma}\right),
\end{equation}
where $\mathbf{z}^{(n)},~n=1,2,...,N$ is the samples, and $\varsigma$ denotes a smooth parameter that corresponds to the width of the window function. Also, the window function $\Pi(\mathbf{z})$ should be a PDF, in order to guarantee its estimated $p(\mathbf{z})$ as a PDF \cite{duda2012pattern}.

In the context of the MCvD applications, as we illustrate from Eq. (12) that the $d$-dimensional metric $\mathbf{z}_k$ follows a Gaussian distribution, it is intuitive to adopt a Gaussian Parzen window to approximate the likelihood PDF $\varphi(\mathbf{z}_k|\alpha_k)$ in Eq. (13). In this view, we assign $\Pi(\mathbf{z})$ as:
\begin{equation}
\Pi(\mathbf{z})=\frac{1}{(2\pi)^{d/2}}\exp\left(-\frac{1}{2}\mathbf{z}^T\mathbf{z}\right).
\end{equation}
As such, conditioned on the different $\alpha_k=j$, the $\mathbf{z}^{(n)}$ sampled within the window $\Pi(\mathbf{z})$ will have the same stochastic properties such that $\lim_{N\rightarrow\infty}1/N\sum_{n=1}^N\mathbf{z}^{(n)}=\bm{\mu}_j$, and $\lim_{N\rightarrow\infty}1/N\sum_{n=1}^N(\mathbf{z}^{(n)}-\bm{\mu}_j)(\mathbf{z}^{(n)}-\bm{\mu}_j)^T=\bm{\Sigma}$.

With the help of Eq. (31), the estimated likelihood can be derived as:
\begin{equation}
\begin{aligned}
\varphi&(\mathbf{z}_k|\alpha_k)\\
&\backsimeq\frac{1}{N}\sum_{n=1}^N\frac{1}{(2\pi\varsigma^2)^{d/2}}\exp
\left(-\frac{(\mathbf{z}_k-\mathbf{z}^{(n)})^T(\mathbf{z}_k-\mathbf{z}^{(n)})}{2\varsigma^2}\right).
\end{aligned}
\end{equation}
Note that as we assign a proper $\varsigma$, we have a perfect approximation with $N\rightarrow\infty$. This is because:
\begin{equation}
\begin{aligned}
&\lim_{N\rightarrow\infty}\frac{1}{N}\sum_{n=1}^N\frac{1}{(2\pi\varsigma^2)^{d/2}}\exp
\left(-\frac{(\mathbf{z}_k-\mathbf{z}^{(n)})^T(\mathbf{z}_k-\mathbf{z}^{(n)})}{2\varsigma^2}\right)\\
\overset{(a)}{=}&\int_{\mathbb{R}^d} p(\mathbf{z}^*)\frac{1}{(2\pi\varsigma^2)^{d/2}}\exp
\left(-\frac{(\mathbf{z}_k-\mathbf{z}^*)^T(\mathbf{z}_k-\mathbf{z}^*)}{2\varsigma^2}\right)d\mathbf{z}^*\\
\overset{(b)}{=}&\int_{\mathbb{R}^d} \varphi(\mathbf{z}^*|\alpha_k)\cdot p(\mathbf{z}_k|\mathbf{z}^*)d\mathbf{z}^*\\
\overset{(c)}{=}&\varphi(\mathbf{z}_k|\alpha_k).
\end{aligned}
\end{equation}
In Eq. (33), (a) accounts for the mean expression of $1/(2\pi\varsigma^2)^{d/2}\exp
(-(\mathbf{z}_k-\mathbf{z}^*)^T(\mathbf{z}_k-\mathbf{z}^*)/(2\varsigma^2))$ given the distribution of $\mathbf{z}^*$. (b) holds for the fact that the sampled $\mathbf{z}^*\sim\varphi(\mathbf{z}^*|\alpha_k)$, and the rest follows a Gaussian PDF conditioned on $\mathbf{z}^*$. The result of (c) resorts to the Chapman-Kolmogorov function \cite{jazwinski2007stochastic}, meaning that likelihood estimation is perfect.

\subsection{Parzen probabilistic neural networks}
In order to implement the Parzen window technique to derive Eq. (32), and further compute Eq. (14) for detection, we introduce the Parzen window based probabilistic neural networks \cite{specht1990probabilistic}. Suppose we plan to form a Parzen estimation based on $N$ patterns, each of which is $d$-dimensional and randomly sampled from $C$ classes. In such a case, the Parzen-PNN will consists of $d$ input units (referred as the input layer), linked to each of the $N$ pattern units. Then, each pattern unit contains a trained data, and connects to one and only one of the class from the total $C$ categories.

The realization structure of the Parzen-PNN is illustrated in Fig. 4. From the input layer, the $(d=3)$-dimensional metric $\mathbf{z}_k$ is directly fed and serves as the Parzen-PNN input. The pattern layer is composed of the $N$ training data, which is supported by a pilot sequence via computing its corresponding $\mathbf{z}^{(n)}$ from Eqs. (7)-(9). The output layer contains $C=2$ class representing 0-bit and 1-bit respectively.

\subsubsection{Training process}
The training process of the Parzen-PNN is relatively simple compared with the other neural network architectures, e.g., the back propagation (BP) networks, and the radius basis function networks (RBF) \cite{haykin2008adaptive}. We provide the training algorithm in Algo. 1, which directly computes and assigns the $N$ values of pattern $\mathbf{z}^{(1)},\mathbf{z}^{(2)},...,\mathbf{z}^{(N)}$ into pattern layer, and then maps each pattern to only one class of $0$-bit and $1$-bit via the knowledge of the pilot sequence.

\begin{figure}[!t]
\centering
\includegraphics[width=3.5in]{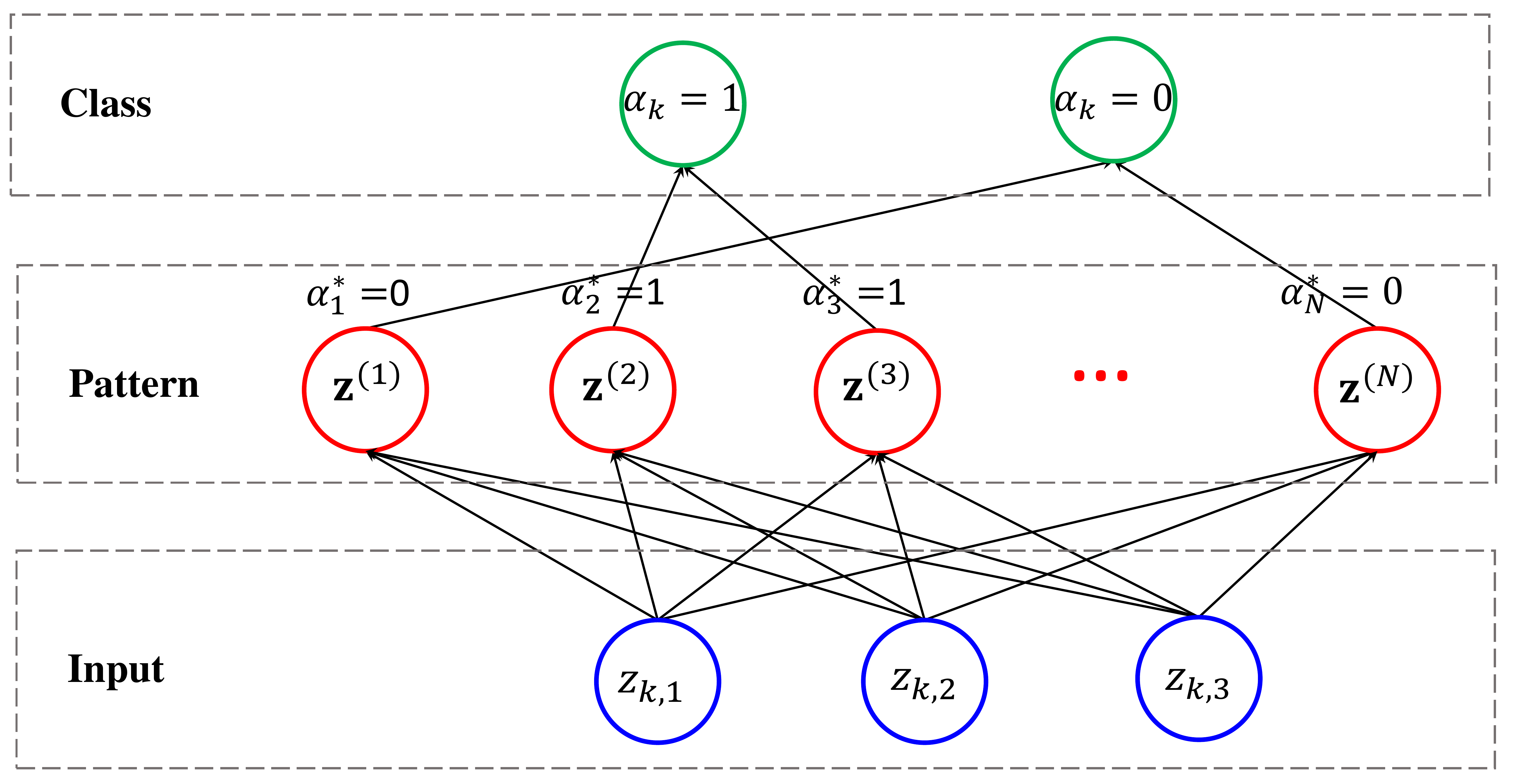}
\caption{Illustration of Parzen-PNN that consists of three layers, i.e., the input layer fed by the designed metric $\mathbf{z}_k$, the pattern layer trained by the pilot sequence $\bm{\alpha}^*=[\alpha_1^*,\alpha_2^*,...,\alpha_N^*]^T$, and the class layer for classification. }
\end{figure}

\subsubsection{Detection process}
After above training process, the Parzen-PNN can be then adopted for signal detection. The detection algorithm is given by Algo. 2. For each time-slot $k$, as we feed the $d$-dimensional metric $\mathbf{z}_k$ into the input layer, each pattern computes the inner difference from its pattern value $\mathbf{z}^{(n)}$, i.e.,
\begin{equation}
T(\mathbf{z}_k,\mathbf{z}^{(n)})=\exp\left(-\frac{(\mathbf{z}_k-\mathbf{z}^{(n)})^T(\mathbf{z}_k-\mathbf{z}^{(n)})}{2\varsigma^2}\right).
\end{equation}
Then, a summation will be computed by adding each $T(\mathbf{z}_k,\mathbf{z}^{(n)})$ in the same category, i.e.,
\begin{equation}
\Delta_{j}(\mathbf{z}_k)=\sum_{\forall n,~s.t.~b_{n,j+1}=1}T(\mathbf{z}_k,\mathbf{z}^{(n)}).
\end{equation}
Finally, the detection can be pursued by the classification decision, as
\begin{equation}
\hat{\alpha}_k=
\begin{cases}
1, &\Delta_1(\mathbf{z}_k)\geq\Delta_0(\mathbf{z}_k),\\
0, &\Delta_1(\mathbf{z}_k)<\Delta_0(\mathbf{z}_k).
\end{cases}
\end{equation}

\begin{algorithm}[t]
\caption{Training Algorithm}
\begin{algorithmic}[1]
\Require
Pilot sequence $\bm{\alpha}^*=[\alpha_1^*,\alpha_2^*,...\alpha_N^*]^T$ and its received data $\mathbf{y}^*=[y_1^*,y_2^*,...,y_{MN}^*]^T$.
\For{$n\in\{1,2,...,N\}$}
\State Compute $\mathbf{z}^{(n)}$ of the pilot $\alpha_n^*$ via Eqs. (7)-(9).
\State Assign $b_{n,j+1}=1$ if $\alpha_n^*=j\in\{0,1\}$.
\EndFor
\Ensure
Return pattern values $\mathbf{z}^{(1)},\mathbf{z}^{(2)},...,\mathbf{z}^{(K)}$, and class indicator $\mathbf{B}=\{b_{n,j+1}\}$.
\end{algorithmic}
\end{algorithm}

\begin{algorithm}[t]
\caption{Detection Algorithm for $k$th symbol}
\begin{algorithmic}[1]
\Require
Received data $\mathbf{y}=[y_{kM},y_{kM+1},...y_{(k+1)M-1}]^T$, and the trained Parzen-PNN.
\State Compute high dimensional metric $\mathbf{z}_k$ from Eqs. (7)-(9).
\For{$n\in\{1,2,...,N\}$}
\State Compute $T(\mathbf{z}_k,\mathbf{z}^{(n)})$ via Eq. (34).
\EndFor
\State Compute $\Delta_1(\mathbf{z}_k)$ and $\Delta_0(\mathbf{z}_k)$ via Eq. (35).
\State Derive $\hat{\alpha}_k$ by Eq. (36).
\Ensure
Return $\hat{\alpha}_k$.
\end{algorithmic}
\end{algorithm}

\subsubsection{Smooth parameter}
As is observed from Eq. (32), the smooth parameter $\varsigma^2$ influences the accuracy of likelihood estimation and thereby the performance of signal detection. To be specific, a small value of $\varsigma^2$ will produce a spiky estimation of the likelihood PDF in Eq. (20), giving rise to a rather curved decision surface and thereby causing extra miss detection. On the other hand, a large value of $\varsigma^2$ will lead to an over-smoothing likelihood density, making it insensitive to the background noise. Here, we give two ideas to assign an appropriate $\varsigma^2$.

In the presence of a known distribution of the channel noise, one can directly compute $\varsigma^2$ with the help of Eq. (12), i.e.,
\begin{equation}
\varsigma^2=|\bm{\Sigma}|^{1/d}.
\end{equation}

Otherwise, if the distribution of the channel noise is unknown, we can still treat the $z_{k,1}$, $z_{k,2}$ and $z_{k,3}$ as Gaussian RVs, according to Eq. (12). Hence, their variances can be estimated via the received data of the pilot sequence, and therefore, Eq. (37) can still be used to compute the proper smooth parameter $\varsigma^2$.

\subsection{Complexity Analysis}
After a complete algorithm description, we investigate the complexities of our Parzen-PNN based high-dimensional non-coherent scheme, by considering both the time and the storage consumptions. In our discussions, we just count the total number of multiplication operations, and treat the exponential operation as several multiplications according to Taylor expansion.

For our designed $(d=3)$-dimensional non-coherent scheme with $K$ input bits, the times of multiplications are $O(5N+5K+dNK)$ that consists of $5N$ for training process, $5K$ for metric construction, and $d\times N\times K$ for the detection process. Given the parallel structure of the Parzen-PNN, the detection process in Eq. (34) can be realized via the $N$ parallel threads, which makes it have only $d\times K$ for time units. Thus, the total time consumption is:
\begin{equation}
\eta(\mathbf{z}_k)=O(5N+5K+dK),
\end{equation}
at the expense of $O(dN)$ storage units. In this view, our scheme is suitable for real time communication, as there are only $O(5\times N)$ for training the PNN, and we spend just $O(d)$ time consumption on detecting each informative bit.

\section{Numerical Simulations}
In the following analysis, the performance of our proposed high-dimensional non-coherent detection scheme will be evaluated, in terms of the BER. First, we examine the performance of its Parzen-PNN realization by considering its smoothing parameter $\varsigma^2$, and the size of the train set $N$ respectively. Then, as the blind detection scenario is considered, the comparisons between our proposed high-dimensional non-coherent scheme, our previous linear non-coherent scheme, and the state-of-the-art MAP are pursued, with respect to various SNR and the bit interval $T_b$ that represents the intensity of the ISI.

The involved parameters in this simulation are configured as follows. As far as both the passive and the absorbing properties of the nano-receiver are concerned, we consider a spherical shape of the receiver with its radius as $R=0.225\mu\text{m}$, and thus its volume as $V=4/3\pi R^3=4.771\times10^{-2}(\mu\text{m})^3$. The sample rate $T_s=M\cdot T_b$ varies from the bit interval $T_b$, but we assure $M=50$ samples for each bit detection. For the diffusive channel, we assign the diffusive coefficient as $D=5\times10^{3}(\mu\text{m})^2/\text{s}$, the communication distance between the transmitter and the receiver as $r=2\mu\text{m}$, the drift velocity as $v=10^{-3}\text{m/s}$, and the degradation factor for enzyme effect as $\beta=100\text{/s}$.

\subsection{Performance of Parzen-PNN}
The performance of the Parzen-PNN realization is illustrated in Fig. 4 and Fig. 5, as we fix the $\text{SNR}=10\text{dB}$, and the $T_b=3\times10^{-4}\text{s}$.

\begin{figure}[!t]
\centering
\includegraphics[width=3.5in]{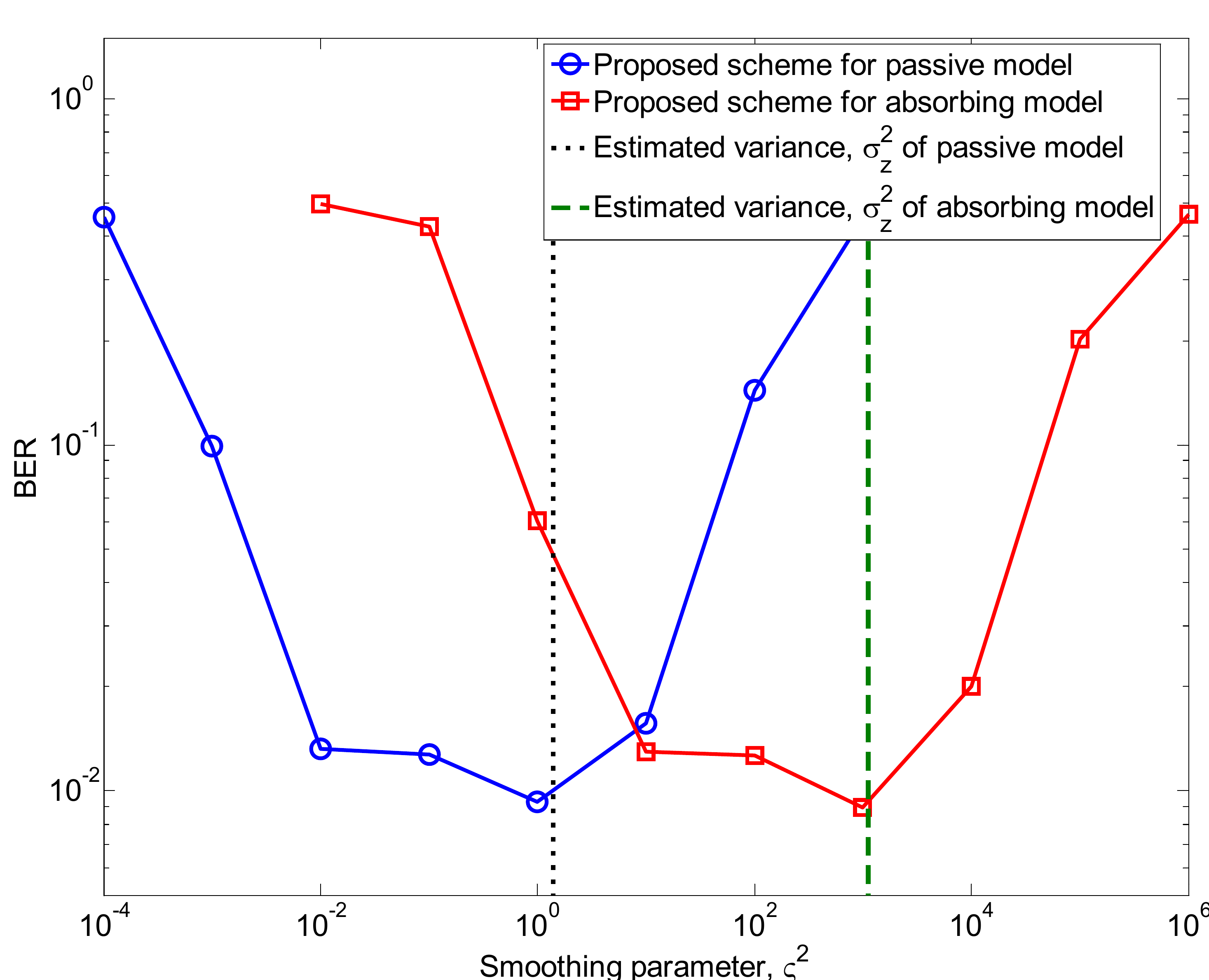}
\caption{Illustration of the importance of the smooth parameter on Parzen approximation and PNN realization of the proposed high-dimensional non-coherent scheme.}
\end{figure}

We can firstly observe from the Fig. 4 that the smoothing parameter $\varsigma^2$ plays an important role on the BER of the Parzen-PNN. Take the passive mode as an example. As $\varsigma^2$ increases from $10^{-4}$ to $10^{4}$, the BER decreases from $0.4$, till reaching its optimal value (i.e., nearly $10^{-2}$), and then begins to rebound. This is because that $\varsigma^2$ reflects the intensity of variance $\bm{\Sigma}$ when approximating the likelihood PDF in Eq. (13) by using Eq. (32), and therefore neither a larger nor a smaller value of $\varsigma^2$ performs well. This can even be demonstrated as we notice that the optimal BER corresponds to the true value of $\varsigma^2=|\bm{\Sigma}|^{1/3}=4.65$, suggesting that a good performance of the Parzen-PNN realization can be achieved when choosing and computing an appropriate smoothing parameter $\varsigma^2$.

Then, we demonstrate that given a proper smoothing parameter, the theoretical performance provided in Eq. (18) can be approached, as we increase the size of the training set, i.e., $N$. It is observed from the Fig. 5 that the size of the training set determines the BER performance of the Parzen-PNN realization. For example, with the increase of the training set from $10$ to $200$, the BERs of both the absorbing and passive models converge to their theoretical limits (i.e., $0.06$ for the absorbing and $0.07$ for the passive), which also verifies the correctness of Eq. (33) that proves the perfect approximations of the Parzen approximation as $N$ grows to infinity.

\begin{figure}[!t]
\centering
\includegraphics[width=3.5in]{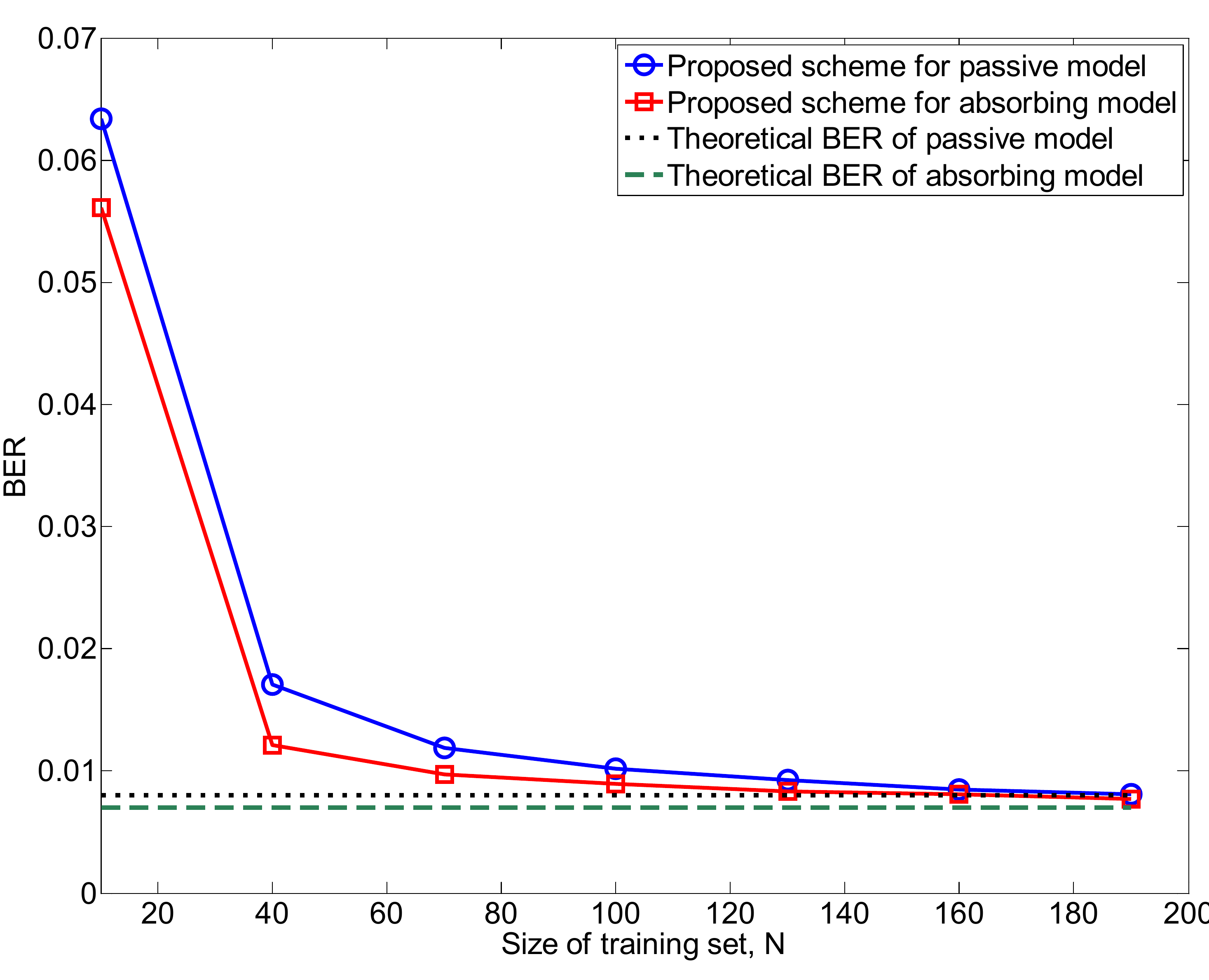}
\caption{Illustration of the BER convergence as the training set grows. }
\end{figure}

With the help of Fig. 4 and Fig. 5, we demonstrate that the Parzen approximation and its PNN realization can approach the theoretical BER performance in the absence of any knowledge of the specific channel models. This indicates that our proposed high-dimensional non-coherent detection scheme is suitable for blind detection scenarios of MCvD, which will be illustrated in the next part.

\subsection{Performance Comparisons}
The BER comparisons between our proposed high-dimensional non-coherent scheme, our previously linear non-coherent scheme, and the state-of-the-art MAP are illustrated in Fig. 6 and Fig. 7, where Fig. 6 gives the comparison with various SNR, and Fig. 7 shows the results affected by the intensities of ISI. Note that, for the two non-coherent schemes, we consider a total unknown channel model, i.e., both the model expressions and the model-related parameters are unknown. For the MAP scheme, we assume a partially unknown condition, i.e., we know that the model is randomly selected from Eq. (2), but the mode for passive or absorbing and all the model-related parameters are unknown, as the MAP cannot compute its likelihood PDF and its transition function if the model formulas are unavailable.

\subsubsection{BER versus SNR}

\begin{figure}[!t]
\centering
\includegraphics[width=3.5in]{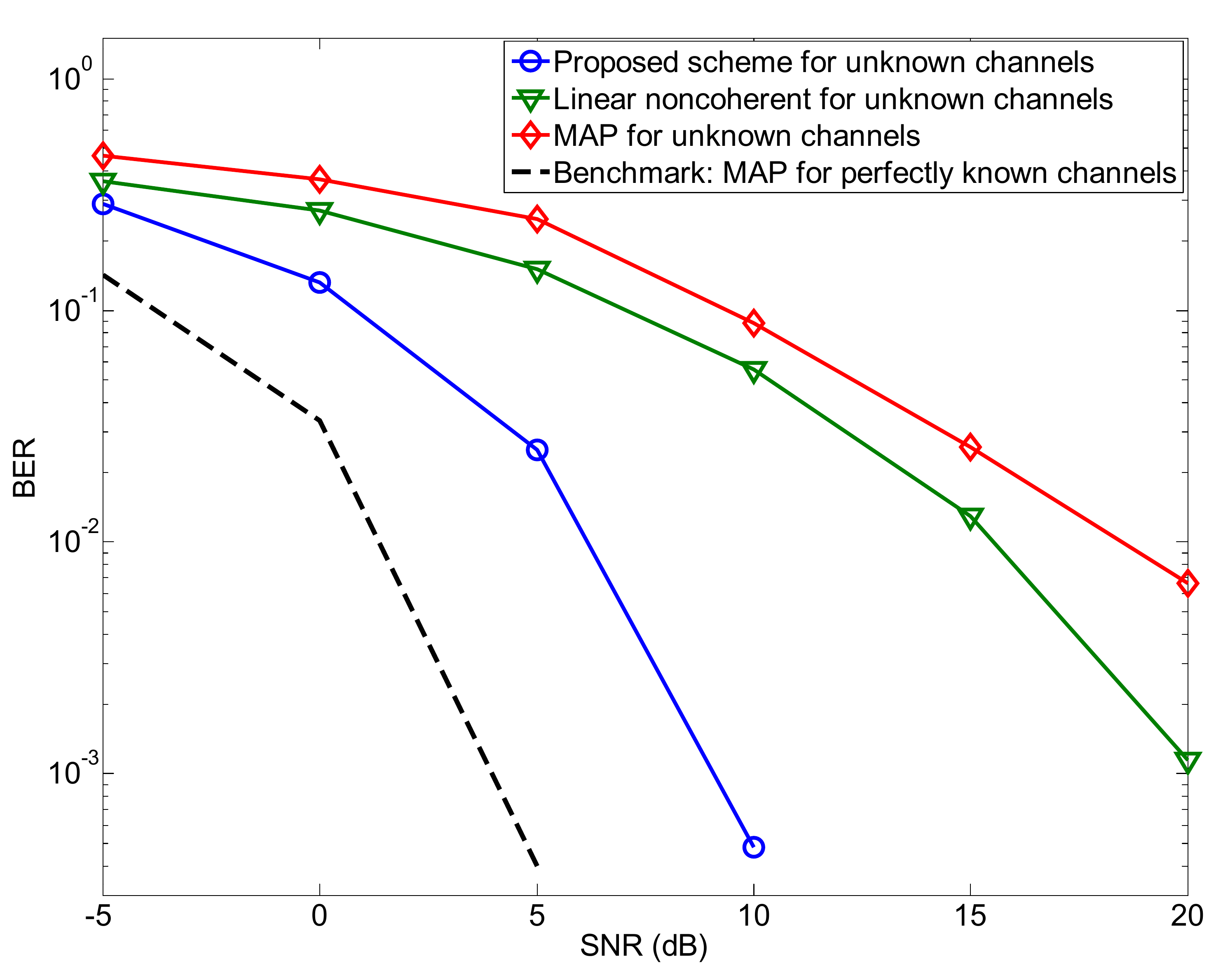}
\caption{BER comparisons among different schemes versus the changes of the SNR.}
\end{figure}

The BER performance with respect to the SNR is shown in Fig. 6. It is intuitive that the BERs of all the schemes are affected by the SNR. For instance, the BER of the MAP is deteriorated from $8\times10^{-3}$ to $0.47$, as the SNR decreases from $20$dB to $-5$dB. Also, our proposed two non-coherent schemes perform badly in the region of the low intensities of the SNR.

Then, it is seen that our proposed high-dimensional non-coherent scheme outperforms the state-of-the-art MAP and our previous linear non-coherent method. For instance, the BER of our high-dimensional non-coherent scheme is $6\times10^{-4}$, as the SNR=$10$dB, which is smaller than that of the MAP scheme (i.e., $0.1$) and the linear scheme (i.e., $8\times10^{-2}$).

The reason can be summarized as two aspects. \textbf{1)} Our proposed scheme relies on the high-dimensional construction of the three feature metrics, rather than their linear combination, capable of providing a larger SNR, and therefore can obtain a better performance given the Ineq. (24). \textbf{2)} Under the assumption of the unknown channel models, the performance of the state-of-the-art MAP scheme is restricted by the accuracy of its estimations of the model parameters, which, if biased, will subsequently deteriorate the computation of the likelihood PDF, therefore leading to erroneous results for signal detection. By contrast, our proposed high-dimensional non-coherent scheme uses the Parzen-PNN, making it possible to approach its theoretical BER bound, and thereby has the ability to reach a remarkable performance in the blind detection scenarios.

\subsubsection{BER versus ISI}

\begin{figure}[!t]
\centering
\includegraphics[width=3.5in]{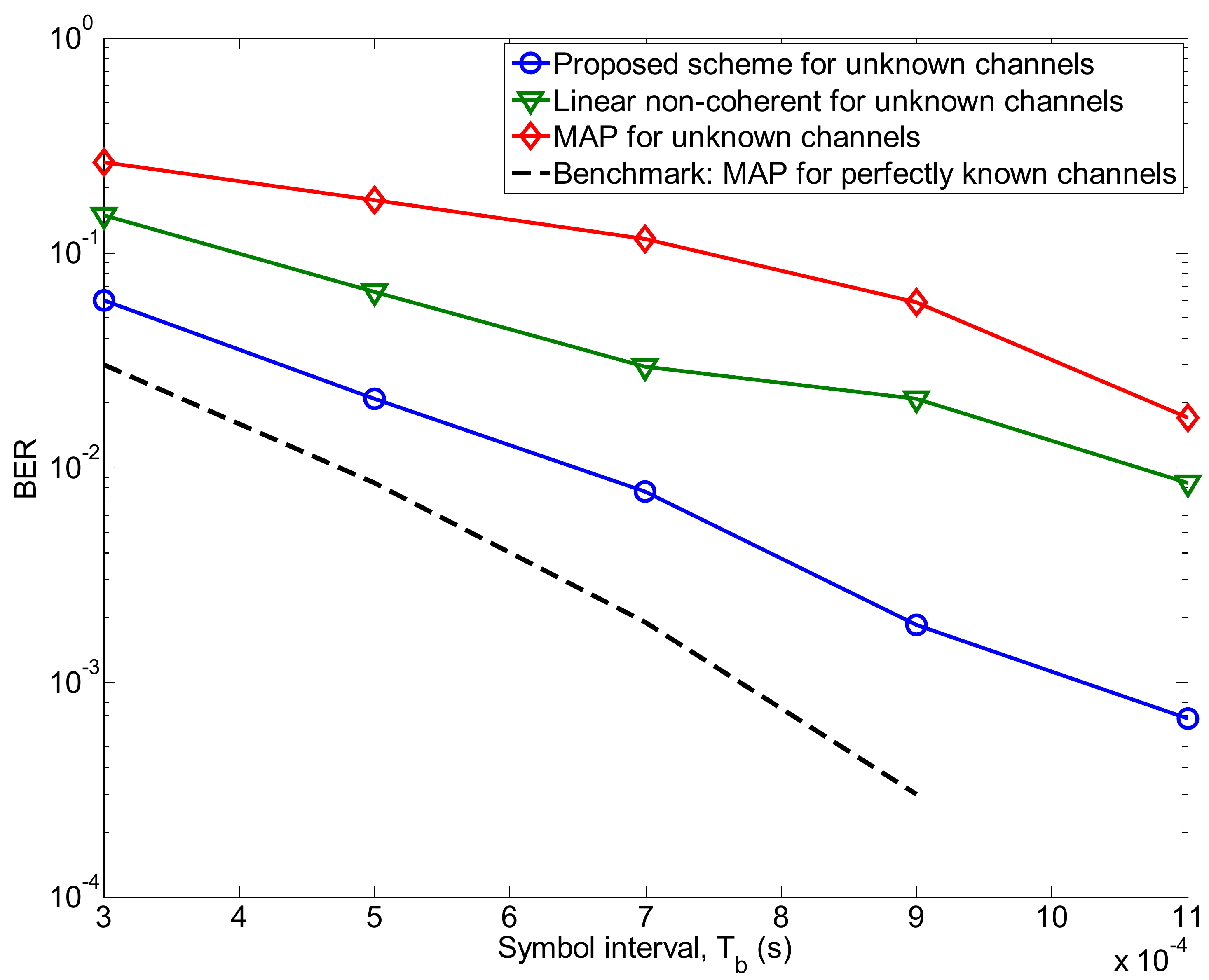}
\caption{BER comparisons among different schemes with respect to the changes of symbol interval $T_b$ that indicate the intensities of ISI.}
\end{figure}

The performance of the BER corresponding to the ISI is provided in Fig. 7. It is shown that the intensities of the ISI influence the BERs of these schemes. For example, with the decrease of the symbol interval $T_b$ from $1.1\times10^{-3}$s to $3\times10^{-4}$, the ISI becomes stronger, and therefore deteriorates the BER of the proposed high-dimensional non-coherent scheme from $9\times10^{-4}$ to $8\times10^{-2}$.

It is noteworthy that in spite of the bad performance from the proposed scheme, it still betters the state-of-the-art MAP and the linear non-coherent scheme. For instance, we can see that the BER of our proposed high-dimensional non-coherent scheme varies from $8\times10^{-2}$ to $9\times10^{-4}$ as the $T_b$ grows from $3\times10^{-4}$ to $1.1\times10^{-3}$, smaller than the values of the linear scheme that ranges from $1.1\times10^{-1}$ to $10^{-2}$, not to mention the poor performances derived from the MAP method.

This great advantage of our proposed scheme is attributed to following perspectives. As we have proved in the Ineq. (24), the high-dimensional non-coherent has a lower BER as opposed to the scheme with lower dimensional metric. Besides, the advantage of blind detection of our scheme makes it possible to approximate the likelihood density and thereby leads to a more reliable detection results, as opposed to the MAP in \cite{6708551} that will be harmed under the unknown channel models.

\section{Conclusion}
In most sequential molecular signal detection scenarios, information recovery without explicit channel models is challenging. Current research challenges lie in how to deal with the background noise the ISI caused by the long-tail nature of the channel response and its dynamics. In this paper, we extend and improve our linear metric combining non-coherent detection method by designing a high-dimensional metric and proposing the Parzen-PNN implementation. We first construct the high-dimensional metric by re-designing the sub-metrics to be robust against various ISI intensities. Then, we compute the theoretical BER bound of the high-dimensional metric and prove to have a lower BER as opposed to the linear scheme. By considering the blind detection scenarios, we adopt the Parzen-PNN implementation to estimate the likelihood PDF and then detect the information bits. In comparison with the state-of-the-art MAP and previous linear schemes, our newly proposed high-dimensional non-coherent scheme gains an average of $10$dB in performance under typical conditions. This generalizable technique provides promising pathways for future research in adverse molecular or biological channels.

\bibliographystyle{IEEEtran}
\bibliography{myref}
\end{document}